\documentclass[letterpaper, 10 pt, conference]{ieeeconf}  

\IEEEoverridecommandlockouts                              

\usepackage{cite}
\usepackage{amsmath,amssymb,amsfonts}
\usepackage{algorithmic}
\usepackage{graphicx}
\usepackage{textcomp}
\usepackage{xcolor}

\title{\LARGE \bf
A Survey on 3D LiDAR Localization for Autonomous Vehicles}

\author{Mahdi Elhousni and Xinming Huang $^{*}$

\thanks{$^{*}$Both authors are with the Electrical and Computer Engineering Department at the Worcester Polytechnic Institute,
        Worcester, MA 01609, USA.
        {\tt\small melhousni@wpi.edu}}%
}

\begin{document}

\maketitle
\thispagestyle{empty}
\pagestyle{empty}

\begin{abstract}
LiDAR sensors are becoming one of the most essential sensors in achieving full autonomy for self driving cars. LiDARs are able to produce rich, dense and precise spatial data, which can tremendously help in localizing and tracking a moving vehicle. In this paper, we review the latest finding in 3D LiDAR localization for autonomous driving cars, and analyse the results obtained by each method, in an effort to guide the research community towards the path that seems to be the most promising.
\end{abstract}

\section{Introduction}
Lately, autonomous driving has become one of the most researched topics in the scientific community. The race toward full autonomous driving cars, or level 6 autonomy such as it categorized by SAE International, is mainly fueled by the hope of eliminating the human errors when it comes to driving. This could have a huge impact on our daily lives : According to a report by The Department of Transportation of the USA, self-driving cars could reduce
traffic fatalities by up to 94\%. \par
In order for a car to drive autonomously, the first challenge that a traditional pipeline would try to solve is to localize the car. Localization in this context means : finding the position and orientation of the vehicle inside of a map. \par 
Defining what a map means is primordial : In this paper, we focus on Autonomous Driving Cars (ADC) with a perception system consistent only of a LIght Detection and Ranging (LiDAR) sensor : a sensor that uses laser light in order to measure distances and is capable of producing PointClouds, which are a 3D representation of space where each point contains the $(x,y,z)$ coordinates of the surface that reflected the laser beam originating from the sensor. The maps used to localize the ADC have to match the available sensor data, which has given rise to PointCoud based maps. These maps can either be constructed beforehand by concatenating successive LiDAR scans offline, or during the navigation process by optimizing and taking advantage of the odometry that could be generated using the LiDAR data and combining that with a loop closure mechanism. \par 
We decided to focus on localization using 3D LiDAR for the following reasons : LiDAR data, compared to other perception sensors, is the richest and most detailed in term of spacial information. This results in the LiDAR sensor being practically always more accurate when it comes to solving spatial based challenges such as localizing a vehicle. We also observed that the price of this sensor is decreasing constantly (from \$75000  a few years ago to less than \$5000  today), making this sensor more accessible to the public and more affordable for car manufacturers. However, localization with 3D LiDARs can face multiple challenges, which usually revolve around efficiency : Since LiDAR sensors tends to produce considerable amounts of data, processing the it in a quick fashion to guarantee the real-time necessities of ADCs can be challenging and usually demands an efficient processing pipeline with some sort of downsampling or feature extraction method. \par
Generating an odometry measurement when localizing a vehicle is an essential step. Multiple approaches have been proposed over the years to utilize the 3D data from LiDARs to calculate the displacement of a subject, which we decided to separate in three distinct categories :\\
\textbf{3D Registration Based Methods :} Usually combined with a map that was built offline, these methods take advantage of the advances that were achieved in 3D pointclouds registration \cite{tam2012registration}. These methods can be seen as ''dense" methods, since they take advantage of all the points present in the LiDAR data.\\ 
\textbf{3D Features Based Methods :} Inspired by the popular methods relying on 2D feature extraction and matching \cite{mur2015orb, mur2017orb, scaramuzza2011visual}, these approaches design relevant features in the 3D space that are then used to calculate the displacements between successive scans. These methods can be seen as "sparse" method, since they only use a select number of points in the LiDAR data. \\ 
\textbf{3D Deep Learning Based Methods :} The use of Deep Learning for solving localization challenges has been gaining more and more popularity lately. 2D camera images were first used to try and predict the odometry between a pair of images \cite{konda2015learning, wang2017deepvo, clark2017vinet, yang2018deep} with results that were more or less acceptable, but still not outperforming the state of the art. Lately, more works have been exploring the use of LiDAR data instead, with results that seem more promising. \par
Results reported by the different publications in these three categories will be compared on the KITTI Odometry Dataset \cite{geiger2013vision}, being the most popular benchmark in the field.
The goal of this survey is to review and present the most relevant work related to 3D LiDAR localization, compare the different results reported in the literature and discuss the strengths and weaknesses of each of them. 

\section{3D LiDAR Localization for Autonomous Driving Cars}

\subsection{3D Registration based methods}
In this section, we review the 3D localization methods based on the 3D point clouds registration approaches. Registration transforms a pair of point clouds in order to align them in the same coordinates frame, making it possible to deduct the transformation between both scans. In the context of ADC localization, registration can be used in two ways : (1) By combining the incoming scans with portions of a pre-built point cloud map in order to localize the vehicle, or (2) By combining successive LiDAR scans in order to calculate the odometry of the vehicle. \par
3D point cloud registration is mostly used in the areas of shape alignment and scene reconstruction, with the Iterative Closest Point (ICP) algorithm \cite{chen1992object, Zhang94iterativepoint} being one of the most popular. In the ICP algorithm, a transformation between a source and target point cloud is iteratively optimized by minimizing an error metric between the points of both point clouds. Multiple variants of the algorithm were developed \cite{rusinkiewicz2001efficient}, such as : point-to-line ICP \cite{censi2008icp}, point-to-plane ICP \cite{low2004linear} and Generalized ICP \cite{segal2009generalized}. The ICP algorithm was the standard for many years to solve point cloud registration tasks, and methods that incorporate it in the localization pipeline were designed, such as in \cite{Mendes2016ICPbasedPS}, where it is combined with a loop closure mechanism and a pose graph building process in order to reduce the accumulated errors from the consecutive registrations. More recently in \cite{kovalenko2019sensor}, an odometry pipeline is proposed which integrates the knowledge about the LiDAR sensor physics and improved the ICP algorithm with a novel downsampling and point matching rejection methods : The downsampling of the LiDAR scans is done using a Normal Covariance Filter (NCF) which keeps only the points with precises normals. On the other hand, the outlier rejection when matching points is achieved with a Geometric Correspondance Rejector : By taking advantage of the rings structure of the LiDAR scans and the normals that were previously calculated, the author defines a threshold named the highest neighbor beam distance, which is used as matching rejection criteria. When plugging both of these method to the ICP algorithm, the author reports a 27\% drop in the drift of the odometry on the KITTI dataset.
\par However, the ICP algorithm was eventually surpassed by the 3D Normal Distribution Transform (NDT) algorithm \cite{magnusson2007scan,magnusson2009three}. First developed to assist autonomous mining vehicles, the 3D NDT is a point cloud registration algorithm that extend the 2D NDT algorithm to the 3D space. Similarly to the ICP algorithm, a transformation between a source and target point cloud is iteratively optimized. But in this case, the error that is being minimized is by first transforming the pointclouds into a probability density functions which can be used with Newton’s algorithm to find the spatial transformation between them. \par
Methods such as the ICP algorithm or the 3D NDT algorithm can produce very accurate transformations between LiDAR pairs, however, in the context of ADC, these methods rarely meet the real-time execution criteria. Also, for these methods to be accurate, an initial guess is usually needed to start the optimization process and avoid local minimas. This usually means having to use an additional sensor (such as an IMU \cite{zhou2017LiDAR, Kuramachi2015GICPSA}) to produce an odometry that could be used as an initial guess, and thus does not fit our LiDAR only setup. \par
In a more classical SLAM approach, the authors in \cite{DBLP:journals/corr/abs-1802-08633} propose a 3 steps algorithm named IMLS-SLAM : First is the dynamic object removal which is simplified to clustering of the scans and removal of small clusters. Second step is to apply a sampling strategy based on the observability of each point in order to down-sample the scan, then finally the matching step where the transformation is optimized by following a scan-to-model matching strategy, using the Implicit Moving Least Square (IMLS) representation. \par
Another popular pre-processing approach that could be applied to the scans before attempting to register them is to compute the surfel (SURFace ELement) representation of the point clouds. In \cite{behley2018efficient}, a surfel map is being built while the incoming scan are being converted into vertex and normal maps that are used to calculate the odometry of the vehicle with a so-called frame-to-model ICP algorithm. The surfel map is then used to find loop closure candidates in order to optimize the trajectory of the vehicle and minimize the drift. An extension to this method was proposed in \cite{Chen2019SuMa}, where a semantic segmentation of a spherical projection of the LiDAR data is used to remove dynamic objects and improve the frame-to-frame matching by enforcing semantic constraints on the frame-to-model ICP algorithm. In \cite{Park2017ProbabilisticSF}, two different surfel representations are computed : an ellipsoid surfel map (ESM) and a disk surfel map (DSM). The ESM, due to it's sparsity, is only used for localization. On the other, the DSM, which is much denser that the ESM is used to reconstruct the surrounding environment. \par
In the same spirit, Collar Line Segments (CLS) construction is a useful pre-processing method that makes it possible to achieve a good level of accuracy when aligning point clouds : In \cite{Velas2016CollarLS}, the LiDAR scans are transformed into line clouds, by sampling line segments between neighbouring points from neighbouring rings. These line clouds are then aligned using an iterative approach : First, the center points of the generated lines are calculated. These points are then used to find the transformation between successive scans by finding the lines in the target pointcloud whose center is closest to the lines in the source pointcloud. Additional post processing is applied to boost the accuracy, using a global optimization based previous transformations. \par
Sometimes, reducing the dimensionality of the LiDAR data can also yield reasonable results, such as in  \cite{sun2018dlo} where the incoming scans are projected onto a 2.5D grid map with occupancy and height. This grid map is equivalent to gray scale image which is used to register the scans based on the photo-metric errors as it is usually done with camera data \cite{Engel-et-al-pami2018}. 
\par

\subsection{3D Features based methods}
In this section, we tackle the 3D localization methods based on 3D features extraction and matching. 3D features \cite{rusu2009fast,Rusu10IROS,steder2010narf, guo2013rotational} are interest points that represent recognizable areas that are consistent in time and space, such as corners and planes. Commonly used for 3D object detection tasks, these features are usually represented using a unique vector called feature descriptor, which can be used to match features in two different point clouds. By finding sufficient and consistent matches, we can calculate the transform between scans using an optimization method and thus construct an odometry measurement. \par
In \cite{Yoneda2014LiDARSF}, the authors propose a study that focuses on finding what type of data and features should be observed when trying to achieve accurate localization of an ADC. The authors here argue that features have to be built and extracted based on the distribution of clusters of points. However, their experiments show that the distribution of points changes drastically from one scene to another, making this method very instable. \par
In an approach named PoseMap, which was proposed in \cite{egger2018posemap}, the authors argue that a 'coherent' map representation of the environment is not necessary to achieve high quality localization : The method takes advantages of a pre-build pointcloud map using \cite{bosse2009continuous}, which is subsampled based on an overlap threshold in order to produce a simple, sparse and light representation of the environment where key poses are maintained. This map representation can be seen as a collection of submaps which can be updated independently from each other, at different points in time. The localization is then solved using a sliding window approach by simply using the two closest submaps to the current vehicle position and minimizing the distance between the old features and the new ones. \par
Geared toward off-road environment, the method proposed in \cite{8500599} and named CPFG-SLAM is inspired by the ICP and NDT algorithms and relies on 3D features and a probability grid map. By taking advantage of the nearest neighbor in the grid instead of the nearest neighbor point, the authors are able to matches and registers point cloud onto the grid map more efficiently. The Expectation-maximization (EM) algorithm is used to estimate the pose, while the final optimization problem is solved using the Levenberg-Marquardt algorithm. \par
Other localization methods try take advantage of predominant geometries that are present in the environment where the ADC will be moving : In \cite{pathak2010online} and \cite{grant2013finding}, plane extraction algorithms are combined with frame-to-frame techniques in order to produce a pose estimation for the vehicle. When compared with the results obtained by the ICP algorithm, plane extraction and alignment methods show great improvement in both accuracy and speed. \par
While being a pure engineering solution that sometimes requires to tweak and adapt multiple parameters, 3D features based localization methods have been known to generate impressive results in both accuracy and speed : Currently holding the first spot in the KITTI odometry leaderboard, the method proposed in \cite{zhang2014loam} starts by extracting planar and corner features based on the smoothness and occlusion of the points. These features are matched with patches of points in the following scan and the Levenberg-Marquardt method is then used to solve the LiDAR motion. As it is usually done in most SLAM pipelines, a map is also being built in the background at a slower frequency than the odometry estimation, which helps with improving the final localization results. An extension to this method was proposed in \cite{shan2018lego} in order to improve its speed and guarantee real-time aspect of the odometry calculation. The main improvements reside in taking advantage of the presence of the ground by removing unreliable features and using a two-step Levenberg-Marquardt method to speed up the optimization step. Still, one of the main remaining issues of the LOAM pipeline is the odometry drift due to the accumulated errors. However, plugging in a loop closure mechanism to the pipeline can solve this issue as it was shown in \cite{unknown} or \cite{8816388}. \par

\subsection{3D Deep Learning based methods}

In this section, we review the 3D localization methods based on deep learning. While still being a very young approach to odometry and localization estimation, the use of deep learning has been gaining more popularity lately after it proved being very promising in the camera domain, and after methods such as PointNet \cite{qi2017pointnet} and PoinetNet++ \cite{qi2017pointnet} showed how efficient deep neural networsk can be when attempting to solve 3D pointclouds related challenges. Usually formulated as a regression problem, methods involving deep learning can either try to solve this task in an end-to-end fashion by using the raw pointclouds as inputs and directly predicting the displacements of the vehicle using a single network, or by trying to substitute certain parts of the pre-established classical pipelines that could benefit for the generalizations possible with deep learning networks. \par
One of the first method that proposed to solve this task using a deep learning approach is \cite{nicolai2016deep} : The idea here is to try and take this challenge back to the image domain instead of attempting to solve it directly in the 3D pointcloud one, in order to simplify the data input of the network. Incoming LiDAR frames are first projected onto the 2D space to produce panoramic depth images which are then fed into a simple 2 branch convolution network, in an attempt to regress the values of the displacement and change in orientation of the vehicle between the two input frames. The results obtained by the authors where subpar when compared with the state of the art. However, they were able to prove that exploring the use of deep learning to solve this task could eventually lead to better results. \par
Panoramic depth images are a popular representation of the LiDAR data, and another method that makes use of them is DeepPCO \cite{wang2019deeppco}. The projected LiDAR frames are fed to a 2 branch network where the first branch predict the translation of the vehicle, while the second one predicts the rotation. \par
Another method that attempts to simplify the input data by projecting onto the 2D space is the one presented in \cite{cho2019deeplo}. Here, the LiDAR frames are projected using the spherical coordinate system to generate two new 2D representations : a vertex map (representing the location $(x,y,z)$ of each point) and a normal map (representing the values of the normals of each point). The proposed network is mainly composed residual blocks and has two major branches : First one, named VertexNet, takes as input the vertex maps and is used to predict the translation between subsequent frames. The second branch, named NormalNet, takes as input the normal maps and is designed to predict the rotation between two subsequent frames. The output of both branches is then combined in order to construct the full transformation between the two LiDAR frames. In order to train the full network in and end-to-end fashion, the authors propose two different training schemes with two different loss function, based on the availability of labeled data : First is a classical supervised loss, where the labeled data is compared with the network prediction in order to optimize the weights of the network, and second is the unsupervised loss, where no labeled data is needed and the ICP algorithm is used to guide the network toward the correct motion predictions. \par
In \cite{yin2018locnet}, and in an effort to simply the input data again, a handcrafted rotational invariant representation (RIR) based on the rings distribution of the point clouds is presented. The authors claim that thanks to this representation, the global localization problem is reformulated to an identity verification one. This was solved by using a siamese network named LocNet, which takes as input 2 subsequent RIR and aims to optimize a contrastive loss function \cite{hadsell2006dimensionality}. The output of LocNet is a dimension reduced feature vector that is used later in complete SLAM pipeline, where the MCL \cite{fox1999monte} and ICP algorithm are used to generate the final transformation in a coarse to fine manner. \par
In \cite{8099748}, the LORAX algorithm was proposed. This approach introduces the notion of super-points, a subset of points located inside of a sphere and describing a local surface, which are projected onto the 2D space to form 2D depth maps. These depth maps are then filtered using a series of test to leave only the relevant super-points and encoded using a PCA and Deep Auto-Encoder. Candidates for matching are then selected based on the euclidien distance between features before engaging in a coarse registration step where an iterative approach involving the RANSAC algorithm is used. As a final step, and in order to fine tune the results of the registration step, the ICP algorithm is used to improve the accuracy of the whole pipeline. \par

\begin{table*}[t]
  \begin{center}
    \caption{Comparison on the KITTI training dataset.}
    \label{tab:table1}
    \resizebox{\linewidth}{!}{%
    \begin{tabular}{|c|c|c|c|c|c|c|c|c|c|c|c|c|}
      \hline
       & \multicolumn{12}{|c|}{Sequences}\\
      \hline
      {Method}  & {00} & {01} & {02} & {03} & {04} & {05} & {06} & {07} & {08} & {09} & {10} & {Avg}\\
      \hline
      G-ICP \cite{segal2009generalized} & 1.29/0.64 & 4.39/0.91 & 2.53/0.77 & 1.68/1.08 & 3.76/1.07 & 1.02/0.54 & 0.92/0.46 & 0.64/0.45 & 1.58/0.75 & 1.97/0.77 & 1.31/0.62 & 1.91/0.73\\
      \hline
      SALO \cite{kovalenko2019sensor} & 0.91/0.72 & 1.13/0.37 & 0.98/0.45 & 1.76/0.50 & 0.51/0.17 & 0.56/0.29 & 0.48/0.13 & 0.83/0.51 & 1.33/1.43 & \textbf{0.64}/\textbf{0.30} & 0.97/0.41 & 0.95/0.80\\
      \hline
      CLS-SLAM \cite{Velas2016CollarLS} & 2.11/0.95 & 4.22/1.05 & 2.29/0.86 & 1.63/1.09 & 1.59/0.71 & 1.98/0.92 & 0.92/0.46 & 1.04/0.73 & 2.14/1.05 & 1.95/0.92 & 3.46/1.28 & 2.13/0.82\\
      \hline
      SuMa \cite{behley2018efficient} & 0.3/0.7 & 0.5/1.7 & 0.4/1.1 & 0.5/0.7 & 0.3/0.4 & 0.2/0.5 & 0.2/0.4 & 0.3/0.4 & 0.4/1.0 & 0.3/0.5 & 0.3/0.7 & 0.3/0.7\\
      \hline
      SUMA++ \cite{Chen2019SuMa} & \textbf{0.22}/0.64 & 0.46/1.60 & 0.37/1.00 & 0.46/0.67 & 0.26/0.37 & 0.20/0.40 & 0.21/0.46 & 0.19/0.34 & 0.35/1.10 & 0.23/0.47 & 0.28/0.66 & 0.29/0.70\\
      \hline
      IMLS-SLAM \cite{DBLP:journals/corr/abs-1802-08633} & -/0.50 & -/0.82 & -/0.53 & -/0.68 & -/0.33 & -/0.32 & -/0.33 & -/0.33 & -/0.80 & -/0.55 & -/0.53 & -/0.55\\
      \hline  
      \hline   
      LOAM \cite{zhang2014loam} & 0.78/0.53 & 1.43/0.55 & 0.92/0.55 & 0.86/0.65 & 0.71/0.50 & 0.57/0.38 & 0.65/0.39 & 0.63/0.50 & 1.12/0.44 & 0.77/0.48 & 0.79/0.57 & 0.85/0.51\\
      \hline
      \hline
      LO-Net \cite{li2019net} & 1.47/0.72 & 1.36/0.47 & 1.52/0.71 & 1.03/0.66 & 0.51/0.65 & 1.04/0.69 & 0.71/0.50 & 1.70/0.89 & 2.12/0.77 & 1.37/0.58 & 1.80/0.93 & 1.09/0.63\\
      \hline      
      DeepLO \cite{cho2019deeplo} & 0.32/\textbf{0.12} & \textbf{0.16}/\textbf{0.05} & \textbf{0.15}/\textbf{0.05} & \textbf{0.04}/\textbf{0.01} & \textbf{0.01}/\textbf{0.01} & \textbf{0.11}/\textbf{0.07} & \textbf{0.03}/\textbf{0.07} & \textbf{0.08}/\textbf{0.05} & \textbf{0.09}/\textbf{0.04} & 13.35/4.45 & 5.83/3.53 & 1.83/0.76\\
      \hline
      DeepPCO \cite{wang2019deeppco} & -/- & -/- & -/- & -/- & 0.02/0.03 & -/- & -/- & -/- & -/- & -/- & \textbf{0.02}/\textbf{0.06} & -/-\\
      \hline
      DeepICP \cite{lu2019deepicp} & -/- & -/- & -/- & -/- & -/- & -/- & -/- & -/- & -/- & -/- & -/- & \textbf{0.071}/\textbf{0.164}\\
      \hline     
      BIAS-COR \cite{tang2018learning} & -/1.42 & -/1.96 & -/0.86 & -/0.83 & -/0.48 & -/0.53 & -/0.41 & -/0.75 & -/1.00 & -/1.00 & -/1.35 & -/1.02\\
      \hline

   \end{tabular}
    }
  \end{center}
\end{table*}

\begin{table}[h]
  \begin{center}
    \caption{Comparison on the KITTI test dataset.}
    \label{tab:table1}
    \resizebox{\linewidth}{!}{%
    \begin{tabular}{|c|c|c|c|}
      \hline
      {Method}  & {Translation Error (m)} & {Rotation Error (deg/m)} & {Runtime (s)}\\
      \hline
      IMLS-SLAM \cite{DBLP:journals/corr/abs-1802-08633} & 0.69 & 0.0018 & 1.25s\\
      \hline
      CPFG-SLAM \cite{8500599} & 0.87 & 0.0025 & \textbf{0.03}\\
      \hline
      SuMa++ \cite{Chen2019SuMa} & 1.06 & 0.0034 & 0.1\\
      \hline
      SuMa \cite{behley2018efficient} & 1.39 & 0.0034 & 0.1\\
      \hline      
      \hline
      LOAM \cite{zhang2014loam} & \textbf{0.55} & \textbf{0.0013} & 0.1\\
      \hline
      \hline
      CAE-LO \cite{yin2020cae} & 0.86 & 0.0025 & 2\\
      \hline
      
   \end{tabular}
   }
  \end{center}
\end{table}

In a series of papers \cite{inproceedings, cramariuc2018learning, dube2018incremental} that eventually led to the final 3D SegMap method \cite{dube2018segmap}, the authors explore how to efficiently extract and encode segments from pointclouds using simple convolution networks, with the hope of solving localization and mapping related tasks. The main contribution of this approach is its data driven 3D segment descriptor which is extracted using a network composed of a series of convolutional and fully connected layers. The descriptor extractor network is trained using a loss function composed of two parts : a classification loss and reconstruction one. Finally the extracted segments and their candidate correspondences are found using the kNearest Neighbors (k-NN) algorithm, which makes it possible to solve the localization task. Note that the 3D SegMap descriptor is a versatile descriptor which can also be used to solve other tasks such object classification. \par
Most of the method discussed previously will inevitably suffer from the presence of dynamic objects (cars, pedestrians ...etc.) in the scene when trying to regress the motion between two frames. Removing dynamic object in the scene has been known to improve the odometry results in most SLAM pipelines. However, detecting then deleting the dynamic objects from the scene in a supervised manner introduces an extra level of complexity which could lead to higher processing times and unstable results. In order to solve this issues in an unsupervised manner, the authors in \cite{li2019net} have proposed to train an encoder-decoder branch for the task of dynamic mask prediction. This is done by optimizing geometric consistency loss function, which indicates areas where geometric consistency can be modeled thanks to the normals of the point cloud data. The full network (named LO-Net) can be trained in an end-to-end fashion by combining the geometric consistency loss, the odometry regression loss and a cross-entropy loss for regularization purposes.  \par
Rather that learning how to localize a vehicle using the LiDAR frames directly, other methods attempt to learn the error model of a classical pipeline. In other words, deep learning can be used to correct the odometry measurements already available, resulting in a powerful and flexible plug in module. The authors in \cite{tang2018learning} have proposed to learn a bias correction term, aimed to improve the results of a classical state estimator that takes LiDAR data as input. The Gaussian Process Model was used to model the 6 odometry errors independently from each other with carefully selected input features that concentrated on the 3 DoF that are most affected by the errors. \par

In \cite{lu2019l3}, a more advanced method named L3-Net was proposed, which can be linked to the bias correction theme, since instead of predicting the full transformation between frames, the authors here are proposing a network that attempts to learn the residual value between their traditional localization system and the ground truth. Relevant features are first being extracted and fed into a miniPointNet to generate their corresponding feature descriptors. A cost volume is then constructed in the solution space $(x, y, z)$ and regularized with 3D convolutional neural networks. Additionally, an RNN branch is added to the network structured to guarantee the temporal smoothness of the displacements predictions. \par

A more complete and general variant of the L3-Net was proposed by the same authors in \cite{lu2019deepicp,lu2019deepvcp} and named DeepICP. Here, the features are being extracted using PointNet++, then filterd using a weighting layer that only keeps the most relevant ones. Similarly to the previous method, the features descriptors are computed using a miniPointNet structure then fed into a corresponding point generation layer, which generate the corresponding key points in the target point cloud. In order to regress the final value of the transformation, two loss function are combined, hoping to encode both the local similarities and global geometric constraints.

\section{Evaluation and discussion}

We compare the previously cited methods based on their reported results on the KITTI odometry benchmark \cite{geiger2013vision}, which is one of the most popular large scale dataset for outdoor odometry evaluation : It contains 22 sequences recorded using a Velodyne HDL-64E that was mounted on top of a car, with LiDAR scans that were already pre-processed to compensate for the motion of the vehicle. Ground truth is available for the 11 first sequences and was obtained using an advanced GPS/INS system. \par
Table II lists all the reported results on the training dataset,
while Table III lists all the results on the test dataset that
were reported on the KITTI official leaderbord. Note that
we only consider the results that do not involve any loop
closure mechanims. While LOAM still occupies the first
position of the KITTI’s leaderboard, it is clear from that
the methods involving deep learning are becoming more and
more accurate. As an exemple, DeepICP’s reported average
result outperform any other proposed method on the training
dataset. However, It is hard for us to qualify them as ”state
of the art” methods for two main reason : (1) DeepICP is
reporting that it takes around 2 seconds to register each
pair of frames. This is too slow to be deloyed on a real
autonomous driving car operating in real life situations, (2)
The results of these approaches on the test dataset have not
yet been reported. Good results on the test dataset would
prove that these method are capable of being used in real
scenarios, and not only on data that the deep neural networks
have already seen. Until then, LOAM and its variant remain
the best option and most trustworthy for real autonomous
driving deployment.

\section{Conclusion}

In this paper, we reviewed, analyzed, compared and
discussed most of the recent advances and findings in
the area of 3D LiDAR localization for autonomous
driving vehicles. We considered systems where the only
sensor used was a 3D LiDAR due to the increasing
importance of this sensor in most accurate perception
and localization systems nowdays, in addition to the
increase in its availability to the general public and
manufacturers. The results on the KITTI odometry
dataset reported by the cited papers where compiled
and compared, leading us to the following statement:
While deep learning based methods are shown to be
producing very promising results and seem to represent
the right path to follow in order to solve this challenge in
the future, methods based on 3D feature detection and
matching are still considered as state of the art due to
their proven stability when deployed in real life scenarios.

{
\bibliographystyle{ieee_fullname.bst}
\bibliography{egbib.bib}
}

\end{document}